\documentclass{article}
\usepackage{enumerate}
\usepackage{titling}
\usepackage{fullpage}
\usepackage{amsmath}
\usepackage{amsfonts}
\usepackage{amsthm}
\usepackage{float}
\usepackage{stmaryrd}
\usepackage{graphicx}
\usepackage{caption}
\usepackage{subcaption}
\usepackage{color}
\usepackage{multicol}
\usepackage{wrapfig}
\usepackage{enumitem}
\usepackage[labelfont=bf]{caption}
\usepackage{mdframed}
\usepackage{framed}
\usepackage[usenames, dvipsnames]{xcolor}
\usepackage{caption}
\usepackage{array}

\newcommand*\samethanks[1][\value{footnote}]{\footnotemark[#1]}

\graphicspath{ {./} }
\newcommand\tab[1][0.5cm]{\hspace*{#1}}
\title{Examining the Use of Neural Networks for Feature Extraction: A Comparative Analysis using Deep Learning, Support Vector Machines, and K-Nearest Neighbor Classifiers}
\author{\textbf{Stephen Notley}\thanks{Department of Computer Science, Rensselaer Polytechnic Institute, Troy, NY} \\notley.sa@gmail.com  \and \textbf{Malik Magdon-Ismail}\samethanks \\ magdon@cs.rpi.edu}
\date{}
\begin{document}

\maketitle

\begin{multicols*}{2}

\section*{Abstract}

\tab Neural networks in many varieties are touted as very powerful machine learning tools because of their ability to distill large amounts of information from different forms of data, extracting complex features and enabling powerful classification abilities. In this study, we use neural networks to extract features from both images and numeric data and use these extracted features as inputs for other machine learning models, namely support vector machines (SVMs) and k-nearest neighbor classifiers (KNNs), in order to see if neural-network-extracted features enhance the capabilities of these models. We tested 7 different neural network architectures in this manner, 4 for images and 3 for numeric data, training each for varying lengths of time and then comparing the results of the neural network independently to those of an SVM and KNN on the data, and finally comparing these results to models of SVM and KNN trained using features extracted via the neural network architecture. This process was repeated on 3 different image datasets and 2 different numeric datasets. The results show that, in many cases, the features extracted using the neural network significantly improve the capabilities of SVMs and KNNs compared to running these algorithms on the raw features, and in some cases also surpass the performance of the neural network alone. This in turn suggests that it may be a reasonable practice to use neural networks as a means to extract features for classification by other machine learning models for some datasets.

\section*{Introduction}

\tab Deep neural networks have found success in a wide variety of applications in modern technology and research. These networks typically consist of many layers, which are responsible for feature extraction, and are terminated by a softmax layer, which does the actual classification. This gives neural networks the unique ability to train feature extraction instead of leaving feature definition to predetermination, as noted in \cite{lecunsvmcnn}. This study explores the idea of replacing the usual softmax classifier in a neural network with a support vector machine (SVM) or k-nearest neighbor (KNN) classifier. This idea has been explored previously, especially with SVMs and image data \cite{cnnsvm17, deeplearninglinearsvm, lecunsvmcnn}, and so the intent of this study is to explore the effects while varying datasets and data types, as well as the neural network architectures and training lengths in hopes of observing this phenomena under varying circumstances.

This effect was examined in both image and numeric data with architectures built specifically for these different data types. This was done in order to compare the efficacy of using neural networks for feature extraction across varying data in order to see if the observed effects were unique to any particular data type or network architecture. The four architectures used to classify the image datasets are convolutional neural networks (CNNs) with varying depth and numbers of convolution layers. All images were greyscale and of size 28x28. The three architectures used to classify based on numeric data are networks comprised of standard, fully-connected layers. These were tested with 5 datasets, which are summarized in table 1 with additional details provided in the corresponding section.

\begin{table*}
\centering
\begin{tabular}{|c || c | c | c | c | c|}
\hline
Dataset & Type & Instances & Classes & Learning Rate & Momentum\\
\hline\hline
MNIST & Image & 70,000 & 10 & 0.01 & 0.9\\
\hline
Network Graphs & Image & 42,167 & 9 & 0.01 & 0.9\\
\hline
CalTech 101 Silhouettes & Image & 8,671 & 101 & 0.01 & 0.9\\
\hline
Statlog (Shuttle) & Numeric & 58,000 & 7 & 0.0001 & 0.9\\
\hline
Seizure Recognition &  Numeric & 11,500 & 5 & 0.0001 & 0.9\\
\hline
\end{tabular}
\captionof{table}{Brief overview of the datasets used in this study}
\end{table*}

\section*{Overview of Basic Models}

\tab The focus of this paper is to introduce a viable method for learning features using neural network architectures and then training standalone classifiers to learn and classify based on these features. As such, we will give a short description of each of the individual learning models involved in creating these hybrid learning methods, placing an emphasis on the intuition behind them, in order to better understand the logic from which these hybrid learning models were conceived. We will, however, be giving only general overviews of these models so as to not get lost in the well-studied details and mathematics of these widely-used learning models, instead placing our focus on the more novel hybrid models.

\subsection*{Neural Networks and Convolutional Networks}
\tab Neural networks (NNs) and convolutional neural networks (CNNs) are powerful learning models that have found many uses and applications in society. These models operate quite differently from many other learning models and are inspired by biological processes for learning information in humans \cite{munakata, cnnbio}. Traditional neural networks consist of layers of neurons (or units) which are fully connected to the previous and subsequent layers. The output of each neuron is fed to each neuron in the subsequent layer and multiplied by a weight. These weights are unique to specific neuron-neuron connections. The neuron then takes all of these weighted inputs and performs some user-defined function on them in order to obtain the output value for that neuron, and the process continues layer by layer until the end of the network is reached. The final layer, usually called the output layer, generally has one neuron representing each class. This last layer typically performs a softmax and the class represented by the output neuron with the highest value is chosen as the model's prediction. Through the use of the backpropagation method with an update function \cite{munakata} the weights are updated and the network is run again until it produces the correct output for the sample. This is repeated for every sample, with one learning run through all of the samples in the training set usually referred to as an epoch. Typically many epochs are used to train a NN, as this leads to a good, generalizeable set of learned weights. Overfitting in these networks can be serious, but there are many methods to counteract it. The one that has found, perhaps, the most success is called dropout. Dropout works to curb the tendency of these networks to fit noise by randomly dropping some units from each fully-connected layer along with the corresponding connections. This is the tool with which we combat overfitting on fully-connected networks in this study \cite{dropout}.

Convolutional neural networks, which are usually used for image data, are similar in most concepts and general terminology, but take advantage of the concept of locality in images. These architectures consist of convolutional layers which pass a number of filters over the image in order to obtain a series of convolved feature maps, one for each filter \cite{lecunsvmcnn}. CNNs also contain pooling layers to downsample these feature maps. These series of convolution and pooling steps are usually terminated by one or more fully-connected layers and a softmax output layer, which serve the same function as in ordinary neural networks. CNNs use the same backpropagation and update model as NNs, but in CNNs it is the filters that are learned, whereas in NNs the weights between layers of neurons are the objective for learning. Both methods have proven extremely effective in learning for a wide variety of purposes. They are somewhat opposite of most traditional learning models, however, as most models emphasize learning a complex classifier, whereas neural network varieties instead place emphasis on learning complex features. Indeed, all of the layers of the network serve to distill and learn increasingly complicated features, with the exception of the output layer, which serves as a relatively simple classifier based on these features.

\subsection*{Support Vector Machines}
\tab Support vector machines (SVMs) are a type of classification model which aims to find the hyperplane with the largest ``margin" separating data into classes. In the separable case, this reduces the problem to placing a hyperplane (or many hyperplanes) in the space of the data such that the distance between the two closest points in opposite classes to the hyperplane are maximized. Often times, however, data is not separable, and so we must define mathematically a penalty for having a data point on the ``wrong" side of the hyperplane. This is incorporated in the form of the hyperparameter $C$, which defines how much weight we give to misclassifying points, and so a large $C$ heavily penalizes misclassified points and urges hyperplanes which fit the training data more closely, whereas a smaller $C$ term allows for more misclassified points in the hopes that this will result in a more generalizable classifier. As such, the $C$ parameter is our main tool to combat overfitting in support vector machines, and is usually determined through the use of validation, as was the case in this study. In the case that we want a non-linear SVM, we can simply define a kernel function $K$ which transforms the input data into another euclidean space. In this case, we employ an RBF (gaussian) kernel. More information on SVMs, and the particulars of the mathematics discussed here, can be found in \cite{svm}.

\subsection*{K-Nearest Neighbor}
\tab K-nearest neighbor classifiers (KNNs)are relatively simple classification models which use a known dataset to classify new data by polling the $k$ closest data points in the known dataset. The new data point is then classified based on the class with the majority representation among the $k$ nearest neighbors. The intuition behind this is very simple- points of the same class should have similar inherent characteristics, and therefore the features of points of the same class should be sufficiently similar to one another that they are ``near" to other points of the same class. The notion of ``nearness" here referring to locations of points in feature space and the distance between them, which is usually represented by euclidean distance. The higher the $k$ term, therefore, the more resistant the classification is to influence by outlier data points, and so $k$ represents a hyperparameter which can be effectively chosen by validation. These classifiers have the benefit of being very simple and intuitive, but tend to struggle with scaling, as classification requires reviewing all data points for every new data point to be classified. Additionally, the classifier must ``remember" all training data points in order to classify new data points, resulting in a large amount of memory usage in order to store all of the training data along with the classification protocols.

\section*{Methods} 
\tab Each dataset was partitioned into a training set, validation set, and test set before beginning the experimental runs. The sizes of these splits varied between datasets, as will be discussed later. Following this, each architecture was formed using Lasagne \cite{lasagne} and was set to train on the combined training/validation set, and then tested on the test set in order to obtain the test performances of the neural networks alone. In order to obtain the SVM and KNN performance baselines, the multiclass SVC with RBF kernel and k-nearest neighbor classifiers from Scikit-Learn \cite{scikit-learn} were used. Validation was used in order to choose the hyperparameters $C$ and $k$ for the SVM and KNN respectively. This was done by allowing each SVM/KNN to train on the training data with each $C$/$k$, testing each in turn on the validation set. The $C$/$k$ that obtained the best performance was then used as the parameter for retraining the model on the combined training and validation dataset. This final trained model was then given the test data, the results of which were used as the baseline performance for the SVM/KNN. The $C$ parameter was chosen from the set of ${0.01, 0.1, 1, 10, 100, 1000}$ while the k parameter was chosen from the set of ${3, 5, 7, 11, 25, \sqrt{N}}$ where $N$ is the number of data points in the training set. The objective function to minimize during training/validation on all models was the categorical cross-entropy.

In order to test the SVM and KNN with the features derived from the neural net, the corresponding architecture was trained again with only the training data (so as not to skew the validation for hyperparameters later). After training, the softmax layer was removed from the architecture. The training set was then run through this ``feature architecture" in order to obtain the training set features. The SVM and KNN were then trained on these neural network outputs and validated for hyperparameters in the same fashion as when these models were trained on the raw features. In the same manner, the test data was first processed by the feature architecture followed by the model in order to obtain the final accuracy metrics for these hybrid models. For simplicity, and to illustrate the difference between the SVMs and KNNs trained on raw features vs. those trained on the features from the neural network, we will refer to the models in which the SVM and KNN were trained on the neural network features as NN/SVM and NN/KNN respectively.

\section*{Data}
\subsection*{MNIST}

\tab This popular benchmark dataset consists of a 60,000-point training set and 10,000-point test set. Of the training set, 10,000 points were reserved for validation, leaving 50,000 training points. Each point is a 28x28 greyscale image of a handwritten digit (0-9) forming 10 natural classes. This is widely-considered to be a relatively easy dataset for images and a standard benchmark \cite{mnist}.

\subsection*{Network Graphs}

\tab This dataset consists of black-and-white network graphs constructed by combining samples from datasets from \cite{graphs1, graphs2, graphs3, graphs4, graphs5, graphs6, graphs6, graphs7, graphs8, graphs9, graphs10} and categorizing based on the network from which the graph was generated. This dataset consists of 42,167 samples, which we partitioned into a training set of size 25,303, and a test and validation set each of size 8,432. The goal of this classification was to predict from which of the 9 data sources these networks were constructed. The large amount of data but fairly abstract nature of the problem made this a moderately difficult image classification problem.

\subsection*{CalTech 101 Silhouettes}

\tab This dataset is an adaptation of the well-known CalTech 101 dataset, which consists of color images of 101 different classes of object \cite{caltech}. This particular adaptation has converted these images to silhouettes of the object, blacking out the entirety of the target object and converting the remaining background to white. This dataset consists of a training set of 6,407 images, of which 2,307 were reserved for validation leaving 4,100 for training, and a test set containing 2,264 images. This dataset is fairly challenging considering the relatively small amount of data for such a large number of classes, and is made even more difficult in conversion to silhouettes due to the color information lost.

\subsection*{Statlog (Shuttle)}

\tab This dataset from the UCI Machine Learning Repository \cite{uci} contains 9 numeric attributes used to classify into one of 7 classes. There are 58,000 data points in total, which were partitioned into a training set of 29,000 data points, and a test and validation set each consisting of 14,500 data points. 80\% of the samples provided in this dataset belongs to one class, and so this information combined with the relatively large amount of data makes this a fairly easy numeric dataset.

\subsection*{Epileptic Seizure Recognition}

\tab This epileptic seizure dataset, again from the UCI Machine Learning Repository \cite{uci}, contains samples belonging to 5 distinct classes of seizure, indicated by 178 numeric attributes. This dataset contains 11,500 data points in total, which were partitioned into a training set of size 7,500 and a test and validation set each of size 2,000. The large number of attributes and relatively small amount of data makes this a fairly challenging learning problem.

\section*{Architectures}

\subsection*{Image Classification Architectures}

\tab The 4 image classification architectures used in the study are convolutional neural networks. CNNs operate by passing filters over image inputs and convolving the images with these filters in order to extract feature maps \cite{lecunsvmcnn}. To visualize this, architecture 1 is pictured in figure 1. CNNs have been found to be effective for classifying images as the use of a moving filter over the image to extract features allows the network to exploit information about locality and edges between figures in images, which tends to be important for accurate classification.

\begin{figure*}
 \centering
 \includegraphics[width=.8\linewidth]{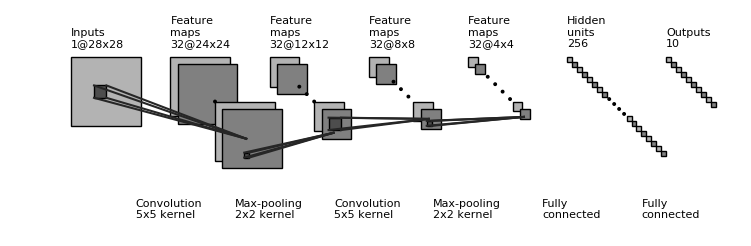}
 \captionof{figure}{Graphical representation of architecture 1 as a general CNN architecture.}
\end{figure*}

\begin{center}
\begin{tabular}{|c || c | c | c| }
\hline
Arch. No. & Conv. Layers & Pool Layers & FC Layers\\
\hline \hline
1 & 2 & 2 & 1\\
\hline
2 & 1 & 1 & 1\\
\hline
3 & 1 & 0 & 1\\
\hline
4 & 0 & 0 & 1\\
\hline
\end{tabular}
\captionof{table}{The numbers of various layer types in the 4 image classification architectures}
\end{center}

The various CNN architectures used in this study explore effects of varying numbers of convolution/pooling steps in the architecture. The study observes how these variations effect the performance of the base CNNs as well as the ability of these networks to extract features for use in the combined models which are the focus of this paper. The differences in the image classification architectures used can be seen in table 2. For all architectures, each convolutional layer uses 32 filters and a stride of 1. Each pooling layer uses max pooling of size 2x2 with stride 2. The fully-connected layers are all hidden layers with 256 hidden units and a dropout rate of $p=0.5$. In addition to these architectures, the networks used to test the performance of the CNN alone for classification added a fully-connected softmax layer to the ends of these architectures which had a number of units equal to the number of classes in the data and a dropout rate of $p=0.5$. For all of these architectures, the learning rate was set to 0.01 and the momentum was set to 0.9.

\subsection*{Numeric Classification Architectures}

\tab In addition to the CNNs meant for image-based inputs, this study also examined 3 standard neural network architectures for classifying numeric data. These architectures contain only fully-connected layers and therefore abandon anything that weighs locality as in the CNNs. These architectures vary in depth in order to examine how the number of fully-connected hidden layers effects classification accuracy for both the neural networks alone, and for the combined models using these networks to extract features. The differences in depth can be seen in table 3. For all of these architectures, the learning rate was set to 0.0001 and the momentum was set to 0.9.

\begin{center}
\begin{tabular}{|c || c | c | c| }
\hline
Arch. No. & FC Layers & Units/Layer & Dropout\\
\hline \hline
5 & 2 & 256 & $p=0.5$\\
\hline
6 & 4 & 256 & $p=0.5$\\
\hline
7 & 6 & 256 & $p=0.5$\\
\hline
\end{tabular}
\captionof{table}{The numbers of various layer types in the 3 numeric classification architectures}
\end{center}

\section*{Results}

\subsection*{Image Data}

\begin{figure*}
 \includegraphics[width=.5\linewidth]{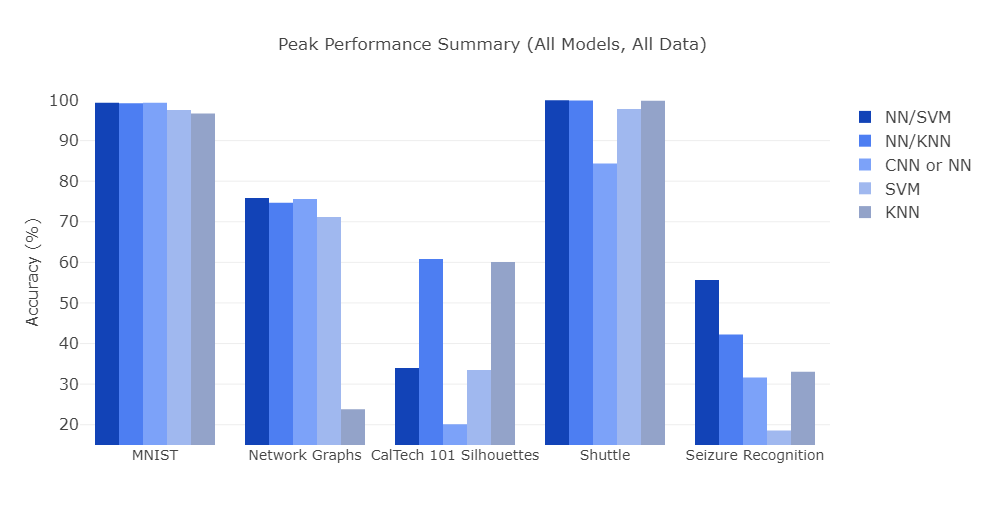}
 \includegraphics[width=.5\linewidth]{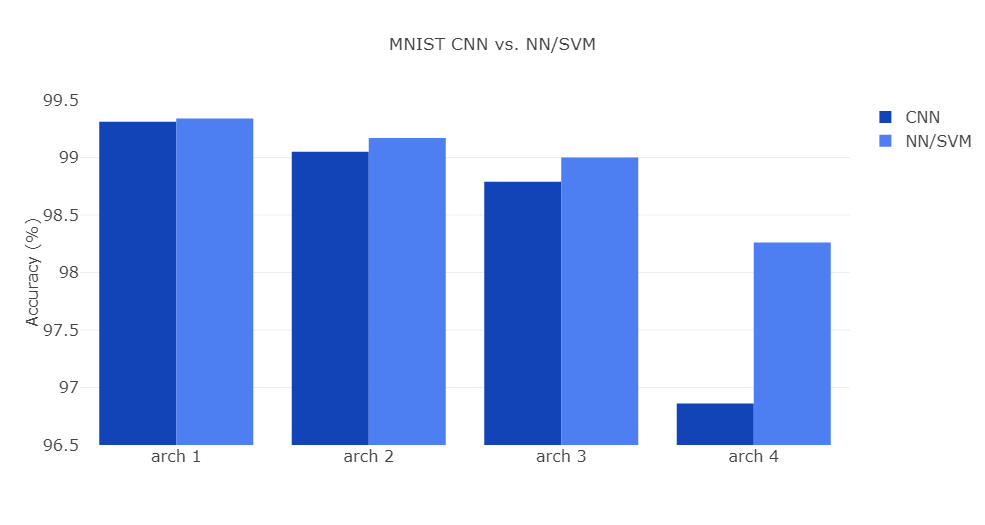}
 \captionof{figure}{Left: Summary of every model's performance across all dataset (highest accuracy across all architectures) plotted using Plotly \cite{plotly}, Right: A close up on the performance on MNIST for all 4 CNN architectures plotted using Plotly \cite{plotly}}
\end{figure*}

\tab On the MNIST dataset, every NN/SVM model performed better than their corresponding CNN-only model. The difference, while small, (on the order of 0.01\%) does show that these NN features do enable higher performance from an SVM as compared to using either a CNN or SVM alone, as can be seen in figure 2. Although the NN/KNN model did not outperform the CNN, it still offered a significant improvement over the KNN run on the raw features.


On the Network Graphs dataset, the NN/SVM again yields the best performance, with 3 of the 4 architectures having superior accuracy when compared to the corresponding CNN architectures. Although the NN/KNN again slightly underperformed when compared to the CNN alone, the NN/KNN performed much better than the KNN alone, offering over 40\% improvement in accuracy compared to the KNN. For the most part, the same trends were observed in the results of testing the Network Graphs dataset as were observed with the MNIST dataset.

The CalTech 101 Silhouettes proved to be a very difficult classification problem for all models in the study. All architectures of CNN performed very poorly, with the NN/SVM and regular SVM performing slightly better and the NN/KNN and KNN alone offering the highest accuracies. Although the NN/SVM and NN/KNN still performed better than their counterparts that trained on the raw features at peak performance, overall the accuracies of these neural-network-trained models hovered very close to their raw-feature-trained versions. This seems to imply that when the neural networks on their own do not perform well and seem to fail to distill meaningful features from the data, this impact carries over to the SVM and KNN trained on these features, giving them accuracies with little to no improvement over the SVM and KNN alone. This result, while somewhat intuitive, is an important pattern to observe.


\subsection*{Numeric Data}

\tab In the Shuttle dataset, all three NN architectures seemed unable to surpass the baseline accuracy of 80\% (recall that 80\% of this dataset belongs to a single class) by more than about 5\%. Despite this, the NN/SVM performed about 2\% better than the SVM alone, while the NN/KNN presented a more modest gain of 0.06\% accuracy at peak performance over the KNN alone. In this dataset, however, the performance of the combined models did not seem to trend upward with the performance of the associated neural network models as we observed in the other datasets. Although the best performance overall still came from the NN/SVM, the NN/SVM and NN/KNN did seem to have as close of a relationship to the performance of their neural-network-only counterparts.

The Epileptic Seizure Recognition dataset proved to be the more difficult numeric dataset. Many of the same observations from the Shuttle dataset were observed again here. We observe the same looser trending in this dataset as was seen in the Shuttle dataset, which contrasts with what was observed in most all instances of the image datasets. We still do see the NN/SVM and NN/KNN outperforming both the regular SVM and KNN as well as the neural network alone, however.

\subsection*{Summary}

\tab The results of testing these various models and architectures against a variety of different input datasets showed that using neural networks for feature extraction is a viable technique for producing strong learning models. The NN/SVM model had the highest peak performance (highest accuracy across all architectures) on every dataset except for the CalTech 101 Silhouettes, on which NN/KNN had the best performance, as can be seen in figure 2 and table 4.

\begin{table*}
\centering
\begin{tabular}{| c ||*{5}{>{\centering\arraybackslash}p{.15\linewidth}|}}
\hline
Model & MNIST & Network Graphs & CalTech 101 Silhouettes & Statlog (Shuttle) & Seizure Recognition\\
\hline \hline
CNN or NN & 99.31\% & 75.63\% & 20.10\% & 84.35\% & 31.65\%\\
\hline
SVM & 97.51\% & 71.17\% & 33.46\% & 97.79\% & 18.55\%\\
\hline
KNN & 96.68\% & 23.79\% & 60.08\% & 99.79\% & 33.05\%\\
\hline
NN/SVM & 99.34\% & 75.87\% & 33.98\% & 99.95\% & 55.65\%\\
\hline
NN/KNN & 99.20\% & 74.66\% & 60.81\% & 99.85\% & 42.25\%\\
\hline
\end{tabular}
\captionof{table}{Summary of peak performances across all models and datasets}
\end{table*}

In most cases, the SVM and KNN both performed better when trained on the neural network features than when their respective models were trained only on the raw features. In many cases, the NN/KNN also performed better than their corresponding architectures using only a neural network. This trend seemed to carry over regardless of the depth of the network. The performance of these combined models on images seemed to increase relative to the performance of the regular neural networks, meaning that as different architectures and training epochs show increasing or decreasing performance, we observe the same trending of increasing/decreasing performance in the corresponding combined models. This trend was also somewhat observed in numeric data, but was much less concrete than in the case of the image datasets tested. Furthermore, accuracy did not seem to linearly scale with architecture depth, as different architectures performed better on different datasets, with some even reaching peak accuracy on some of the most shallow architectures.

\section*{Conclusions}

\tab The results from this study seem to indicate that, in many cases, combining features derived from neural networks with alternate classification models can give impressive accuracy results. Often times neural networks are considered to be an all-purpose tool that can be applied to any learning problem to achieve maximum learning performance. The results of this study seem to indicate that the true power of these models are in their capacity to extract powerful and complex features from many different forms of data and, by extension, that it may be worthwhile to consider applying these features to a variety of learning models, rather than using the neural network alone. This allows us to achieve the best of both worlds, so to speak, exploiting the powerful feature extraction abilities of neural networks alongside the ability to learn a complex classifier via these other learning models.

\section*{Related Work}

\tab The findings here agree with the findings in \cite{deeplearninglinearsvm, lecunsvmcnn}, both of which also found at least partial success in using SVMs on features learned from neural networks. The results in \cite{cnnsvm17} showed the SVM with CNN features slightly underperforming, with a CNN alone converging much more tightly over larger numbers of training steps, which suggests an area of future work for this study. There are, however, many instances where learned features with complex classifiers seem to yield at least as good a result as the individual models alone, and so this presents further evidence that this method is suitably powerful to be worth consideration when choosing a model for learning problems.

\section*{Future Work}

\tab As noted previously, in some instances there seems to be indication that the benefit of using a classifier other than the usual softmax classifier in the output layer of the neural network may deteriorate with increasingly large numbers of epochs \cite{cnnsvm17}, so testing these same architectures and datasets across much larger numbers of epochs could provide further interesting insights into this phenomenon. Additionally, exploring the concept of extracting features from pre-trained versions of famous networks, such as LeNet or GoogLeNet \cite{cnn, googlenet} for use with other classifiers in order to assess if any improvement could provide interesting insight.

This study focused on the comparison between features distilled from neural networks and the simple raw features of data in terms of enabling classification ability in machine learning models. To extend this, it could prove interesting to compare features suggested by domain experts of certain datasets to neural network features in order to examine whether or not these more human-informed features would be superior to those derived using the neural network.

Additionally, the NN/KNN model could benefit from  the application of nearest-neighbor condensing methods \cite{surveycknn}. These methods could prune the data down to a small subset of representative data points in order to provide faster classification times and much less memory usage, increasing the practicality of use of this model. This may also, however, result in lower accuracy for the NN/KNN method. Therefore examining the effects of these condensing methods on speed, memory, and accuracy in NN/KNN models and KNN models presents a potentially interesting area for further exploration.

\section*{Acknowledgements}

\tab We would like to thank the Computer Science department at Rensselaer Polytechnic Institute for providing the resources and support that made this study possible.

This research was sponsored by the Army Research Laboratory and was accomplished under Cooperative Agreement Number W911NF-09-2-0053. The views and conclusions contained in this document are those of the authors and should not be interpreted as representing the official policies, either expressed or implied, of the Army Research Laboratory or the U.S. Government. The U.S. Government is authorized to reproduce and distribute reprints for Government purposes notwithstanding any copyright notation here on.

\bibliographystyle{plain}
\bibliography{research}

\begin{thebibliography}{10}

\bibitem{graphs9}
{JJATT}, 2009.
\newblock http://dostapps.jjay.cuny.edu/ jjatt/data.php.

\bibitem{cnnsvm17}
Abien Fred~M. Agarap.
\newblock An architecture combining convolutional neural network (cnn) and
  support vector machine (svm) for image classification.
\newblock Technical report, 2017.

\bibitem{svm}
Christopher~J.C. Burges.
\newblock A tutorial on support vector machines for pattern recognition.
\newblock {\em Data Mining and Knowledge Discovery}, 2:121--167, 1998.

\bibitem{graphs10}
E.~Cho, S.~A. Myers, and J.~Leskovec.
\newblock Friendship and mobility: user movement in location-based networks.
\newblock {\em {KDD}}, 2011.

\bibitem{uci}
Dua Dheeru and Efi Karra~Taniskidou.
\newblock {UCI} machine learning repository, 2017.

\bibitem{lasagne}
Sander Dieleman, Jan Schlüter, Colin Raffel, Eben Olson, Søren~Kaae
  Sønderby, Daniel Nouri, et~al.
\newblock Lasagne: First release., August 2015.

\bibitem{graphs2}
J.~Gehrke, P.~Ginsparg, and J.~Kleinberg.
\newblock Overview of the 2003 {KDD} cup.
\newblock {\em {SIGKDD} Explor. Newsl.}, 2003.

\bibitem{plotly}
Plotly~Technologies Inc.
\newblock Collaborative data science, 2015.

\bibitem{cnnbio}
SR~Kheradpisheh, Masoud Ghodrati, Mohammad Ganjtabesh, and Timothee Masqulier.
\newblock Deep networks can resemble human feed-forward vision in invariant
  object recognition.
\newblock {\em Scientific Reports}, 6, 2016.

\bibitem{cnn}
Yann LeCun, Leon Bottou, Yoshua Bengio, and Patrick Heffner.
\newblock Gradient-based learning applied to document recognition.
\newblock In {\em Proceedings of the {IEEE}}, 1998.

\bibitem{mnist}
Yann LeCun, Cortes Corinna, and Christopher~J.C. Burges.
\newblock The mnist database of handwritten digits.
\newblock http://yann.lecun.com/exdb/mnist/, 2010.

\bibitem{lecunsvmcnn}
Yann LeCun and Fu~Jie Huang.
\newblock Large-scale learning with svm and convolutional nets for generic
  object recognition.
\newblock {\em Computer Vision and Pattern Recognition}, 2:284--291, 2006.

\bibitem{graphs7}
J.~Leskovec, L.~A. Adamic, and B.~A. Huberman.
\newblock The dynamics of viral marketing.
\newblock {\em {TWEB}}, 2007.

\bibitem{graphs4}
J.~Leskovec, Lang~K. J., A.~Dasgupta, and M.~W. Mahoney.
\newblock Community structure in large networks: Natural cluster sizes and the
  absence of large well-defined clusters.
\newblock {\em Internet Mathematics}, 2009.

\bibitem{graphs1}
J.~Leskovec, J.~Kleiberg, and C.~Faloutsos.
\newblock Graphs over time: Densification laws, shrinking diameters and
  possible explanations.
\newblock {\em {KDD}}, 2005.

\bibitem{graphs3}
J.~Leskovec and J.~J. Meauley.
\newblock Learning to discover social circles in ego networks.
\newblock {\em {NIPS}}, 2012.

\bibitem{caltech}
Benjamin~M. Marlin.
\newblock Caltech 101 silhouettes data set.
\newblock https://people.cs.umass.edu/~marlin/data.shtml, 2014.

\bibitem{surveycknn}
Amal Miloud-Aouidate and Ahmed~Riadh Baba-Ali.
\newblock Survey of nearest neighbor condensing techniques.
\newblock {\em (IJASCA) International Journal of Advanced Computer Science and
  Applications}, 2(11):59--64, 2011.

\bibitem{munakata}
Toshinori Munakata.
\newblock {\em Fundamentals of the New Artificial Intelligence: Beyond
  Traditional Paradigms}.
\newblock Springer, 1998.

\bibitem{scikit-learn}
F.~Pedregosa, G.~Varoquaux, A.~Gramfort, V.~Michel, B.~Thirion, O.~Grisel,
  M.~Blondel, P.~Prettenhofer, R.~Weiss, V.~Dubourg, J.~Vanderplas, A.~Passos,
  D.~Cournapeau, M.~Brucher, M.~Perrot, and E.~Duchesnay.
\newblock Scikit-learn: Machine learning in {P}ython.
\newblock {\em Journal of Machine Learning Research}, 12:2825--2830, 2011.

\bibitem{dropout}
Nitish Srivastava, Geoffrey Hinton, Alex Krizhevsky, Ilya Sutskever, and Ruslan
  Salakhutdinov.
\newblock Dropout: A simple way to prevent neural networks from overfitting.
\newblock {\em Journal of Machine Learning Research}, 15:1929--1958, 2014.

\bibitem{googlenet}
Christian Szegedy, Wei Liu, Pierre Sermanet, Scott Reed, et~al.
\newblock Going deeper with convolutions.
\newblock Technical report, 2014.

\bibitem{deeplearninglinearsvm}
Yichuang Tang.
\newblock Deep learning using linear support vector machines.
\newblock Technical report, 2015.

\bibitem{graphs5}
R.~West and J.~Leskovec.
\newblock Human wayfinding in information networks.
\newblock {\em {WWW}}, 2012.

\bibitem{graphs6}
R.~West, J.~Pineau, and D.~Precup.
\newblock Wikispeedia: An online game for inferring semantic distances between
  concepts.
\newblock {\em {IJCAI}}, 2009.

\bibitem{graphs8}
J.~Yang and J.~Leskovec.
\newblock Defining and evaluating network communities based on ground-truth.
\newblock {\em {ICDM}}, 2012.

\end{thebibliography}

\end{multicols*}
\end{document}